\documentclass{article}

\PassOptionsToPackage{numbers}{natbib}

\usepackage[nonatbib]{neurips_2020}

\usepackage[utf8]{inputenc} 
\usepackage[T1]{fontenc}  
\usepackage{hyperref}  
\usepackage{url}            
\usepackage{booktabs}       
\usepackage{amsfonts}      
\usepackage{nicefrac}       
\usepackage{microtype}      
\usepackage{amsmath}
\usepackage{natbib}
\usepackage{multirow}
\usepackage{kotex}
\usepackage{xcolor}

\usepackage{bm,amsmath,amssymb}
\usepackage{xspace}
\usepackage{graphicx}
\usepackage{color}
\usepackage{algorithm}
\usepackage{algorithmic}

\usepackage{colortbl}
\definecolor{light_gray}{RGB}{170,170,170}

\usepackage[utf8]{inputenc} 
\usepackage[T1]{fontenc}  
\usepackage{url}            
\usepackage{booktabs}       
\usepackage{amsfonts}       

\usepackage{nicefrac}       
\usepackage{microtype}     
\usepackage{longtable,lscape}
\usepackage{array,multirow}
\usepackage[bb=boondox,bbscaled=.95]{mathalfa}

\usepackage{subcaption}

\usepackage{wrapfig}

\usepackage{enumitem}

\usepackage{bm}

\usepackage[main=english, vietnamese]{babel}

\title{Evaluation of Out-of-Distribution Detection Performance of Self-Supervised Learning in a Controllable Environment}

\author{Jeonghoon Park$^{1}$\thanks{Equal contribution} , Kyungmin Jo$^{1}$\footnotemark[1] , Daehoon Gwak$^{2}$\footnotemark[1] , Jimin Hong$^{3}$, \\
 \textbf{Jaegul Choo$^{1}$, Edward Choi$^{1}$} \\
 KAIST$^{1}$, Korea University$^{2}$, Humelo$^{3}$ \\
 \texttt{\{jeonghoon\_park, bttkm, jchoo, edwardchoi\}@kaist.ac.kr} \\
 \texttt{hune282@korea.ac.kr, jimin9401@gmail.com}
}

\begin{document}

\maketitle

\begin{abstract}
We evaluate the out-of-distribution (OOD) detection performance of self-supervised learning (SSL) techniques with a new evaluation framework. Unlike the previous evaluation methods, the proposed framework adjusts the distance of OOD samples from the in-distribution samples. We evaluate an extensive combination of OOD detection algorithms on three different implementations of the proposed framework using simulated samples, images, and text. SSL methods consistently demonstrated the improved OOD detection performance in all evaluation settings.
\end{abstract}

\section{Introduction}

\label{sec:intro}

An out-of-distribution (OOD) detection task is to recognize outliers or anomalies, which do not follow the distribution of the training data.
To address this problem, numerous methods have been proposed in classification tasks in several modalities, including computer vision \cite{hendrycks17baseline, lee2018simple, liang2018enhancing, hendrycks2019oe, vyas2018out}, natural language processing \cite{hendrycks2020pretrained, tan-etal-2019-domain}, and two-dimensional real-number dataset (\textit{e.g.,} Gaussian noise distribution) \cite{ren2019likelihood, lee2018simple, padhy2020revisiting, winkens2020contrastive}. Especially, \cite{hendrycks17baseline, lee2018simple, liang2018enhancing} are simple yet efficient algorithms since they use existing trained models for OOD detection without additional fine-tuning with OOD samples.

Recently, the rapid progress of the Self-Supervised Learning (SSL) achieved a significant performance improvement in various tasks of computer vision \cite{grill2020bootstrap, chen2020simple} and natural language processing \cite{devlin-etal-2019-bert, brown2020language}. Also, self-supervised learning has shown its potentials in OOD detection by using contrastive learning \cite{tack2020csi, winkens2020contrastive}, masked language modeling \cite{hendrycks2020pretrained} and rotation prediction \cite{hendrycks2019selfsupervised}.

However, the OOD detection performance of a model, including SSL-based ones, is typically evaluated by the AUROC score of the binary classification between in-distribution (ID) and OOD, which do not appropriately consider the characteristics between ID and OOD samples.
For example, we can reasonably assume that the distance between the pixel distributions of CIFAR10 and CIFAR100 is closer than the distance between CIFAR10 and MNIST.
Therefore, if the distance between ID and OOD correlates with the difficulty of OOD detection \cite{lee2018training}, then, given CIFAR10 as the ID, CIFAR100 can be considered a harder OOD dataset than MNIST. 

The degree of difficulty for OOD detection was considered in some previous studies using maximum mean discrepancy between two datasets or confusion log probability~\cite{liang2018enhancing, winkens2020contrastive}.
Additionally, \cite{hendrycks2019nae} proposed an OOD dataset called ImageNet-O consisting of hard OOD samples with low-level statistics similar to ID (i.e., ImageNet~\cite{imagenet}).
And recently, \cite{winkens2020contrastive,tack2020csi} argued that contrastive loss is effective for detecting harder OOD samples based on the performance using near classes or confusing examples.
Although such existing methods estimate the difficulty of OOD detection to some extent, or introduce a novel dataset to provide a more challenging setup, they cannot adjust the difficulty of OOD detection.
Therefore, a principled evaluation framework which can consider the distance of two distributions (\textit{i.e.,} the difficulty of OOD detection) is required to observe and compare the behavior of OOD detection methods to truly evaluate their performance.

In this paper, we propose a preliminary approach for a controllable OOD detection evaluation framework using datasets with three different modalities (simulated samples, images, and text). In this framework, we evaluate existing OOD detection algorithms, including SSL. Finally, we show that experimental results support the validity of our assumptions for controllable settings, and reveal interesting insights. Additionally, the results in three modalities consistently validate the effectiveness of SSL for OOD detection.

\section{Performance evaluation of SSL in OOD detection}

\label{sec:exp}
In this section, we describe the proposed controllable evaluation framework followed by the detailed experiment settings as well as evaluation results for each modality.

\subsection{Controllable Evaluation Framework}
We propose a controllable evaluation framework for OOD detection tasks and implement it with three different modalities: 2D simulated samples, images, and text documents\footnote{All descriptions and results regarding text is provided in the appendix}. The core idea of this framework is gradually adjusting a distance between ID samples and OOD samples using a distance factor.

\begin{wrapfigure}[12]{R}{0.45\textwidth}
    \begin{center}
        \vspace{-4mm}
        \includegraphics[width=0.45\textwidth]{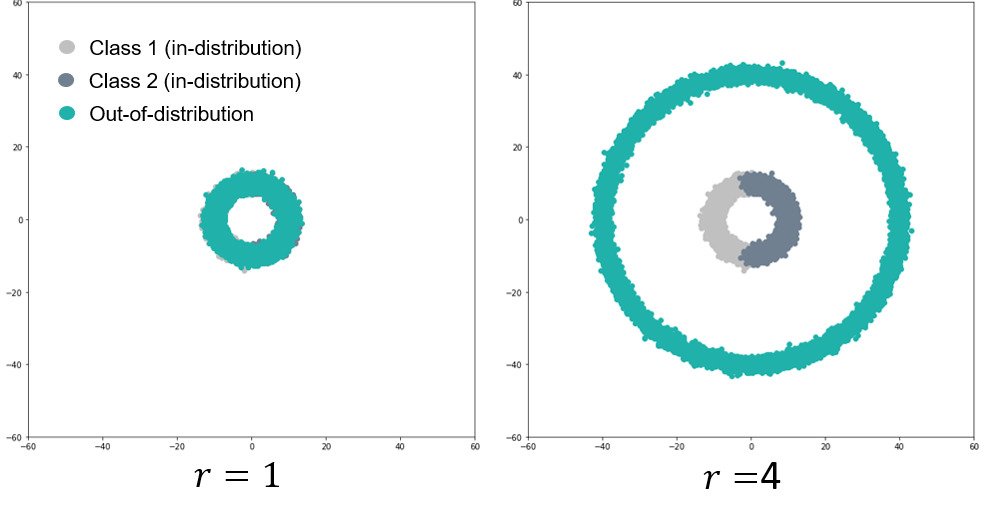}
        \caption{Examples of 2D circle dataset}
        \label{fig:Inputs}
    \end{center}
\end{wrapfigure}

\textbf{2D Simulated Dataset} \enskip A simple simulated dataset was generated in a 2D space to validate the effects of self-supervised methods in OOD detection. Fig.~\ref{fig:Inputs} shows ID samples coming from two classes, and the controllable OOD samples that gradually moves away from the ID samples using the multiplying factor $r$.
OOD samples follow the circle-shaped distribution whose radius is $r$ times larger than the radius of ID samples, with $r=1$ indicating the case that OOD equals ID.
Binary classification between two ID classes is trained with 48,000 samples for training and 16,000 samples for validation.
The OOD detection performance is evaluated using the 16,000 ID test samples and 16,000 OOD samples.
In this experiment, a set of OOD samples with 50 different distances ($1 \le r \le 5.9$) was used to evaluate all OOD detection methods.

\textbf{Images} \enskip In order to build a controllable dataset similar to the 2D circle dataset, Mixup\cite{zhang2018mixup} was used to generate data samples between ID and OOD by adjusting the mixing ratio of OOD to ID images (Fig.~\ref{fig:img_mixup_example} in Appendix.~\ref{appendix:dataset}), under the assumption that different mixing ratios would correlate with the ID-OOD distance.
We use each one of CIFAR10 and CIFAR100 as ID samples while using TinyImagenet, LSUN, SVHN, and CIFAR100 or CIFAR10 as OOD samples. 
For rigorous comparison, all experiments were conducted on the same pre-built mixup dataset consisting of 10,000 samples for each mixing ratio, using 10,000 ID and OOD samples only once for each.

\subsection{Methods}
Same as \cite{hendrycks2019selfsupervised, tack2020csi, winkens2020contrastive}, we train the models using a multitask approach consisting of the supervised classification objective function $L_{cls}$ and self-supervised objective function $L_{ss}$.
Thus, our objective function can be written as $L_{multi}=L_{cls} + \alpha L_{ss}$, where coefficient $\alpha$ is a hyperparameter. 
In order to analyze the behavior, we conduct and evaluate the experiments in several aspects: (1) classification methods, (2) self-supervised methods, and (3) methods for OOD scoring. 

\begin{table}[h]
\vspace{-5mm}
\caption{Classification accuracy(\%) of in-distribution dataset}
\label{table:ImgInDClsfAcc}
\centering
\resizebox{\textwidth}{!}{\begin{tabular}{c c c c c c c} 
\hline\hline 
Dataset &CE &CE+SimCLR  &CE+BYOL &OVADM &OVADM+SimCLR &OVADM+BYOL
\\ [0.1ex]  
\hline   

2D Circle   &97.44  &\underline{97.46}  &97.45  &\textbf{97.49}  &97.45  &97.44
\\ [0.5ex]  

CIFAR10   &94.61	&\underline{94.87}	&\textbf{95.39}	&93.23	&94.67	&94.84
\\ [0.5ex]  

CIFAR100   &76.23	&76.96	&\textbf{78.38}	&73.8	&74.38	&\underline{77.13}
\\ [0.1ex]  

\hline 
\end{tabular}}
\end{table}

\textbf{Classification methods}
For the classification objective function $L_{cls}$, we use a softmax classifier with a cross-entropy loss (CE) or a One-Vs-All classifier with Distance Maximization loss (OVADM) \cite{padhy2020revisiting}. We use the OVADM loss and the CE loss to analyze the effect of distance-based loss in OOD detection.

\textbf{Self-supervised methods}
In the 2D circle dataset and the image datasets, we use the same self-supervised objective functions used in BYOL \cite{grill2020bootstrap} and SimCLR \cite{chen2020simple}. We compare BYOL and SimCLR to compare the effect of using only positive samples when training a self-supervised task in OOD detection.

\textbf{OOD scoring methods}
To detect OOD samples, we use three well-known OOD scoring methods, Baseline (max value of the output vector of the classifier) \cite{hendrycks17baseline}, Mahalanobis Distance(MD) \cite{lee2018simple}, and ODIN \cite{liang2018enhancing}. The Baseline and ODIN are not distance-based, but MD is a distance-based method.

\textbf{Model architecture}
To classify the 2D circle dataset, we used a simple network consisting of two fully-connected layers to obtain features of a penultimate layer.
When training with self-supervised methods, we considered adding noise to the input to emulate the data augmentation of SimCLR and BYOL in a 2D space.
The noise was sampled from a normal distribution with a mean of zero and a standard deviation of 0.01.
For the experiments with image datasets, Wide ResNet \cite{zagoruyko2016wide} with the depth of 28 and the width of 10 was used.
The data augmentation technique of the original SimCLR paper was applied equally to all experiments except for resizing and crop.
The other training details follow the original Wide ResNet paper.

\subsection{Results}    

We performed experiments in three different modalities and evaluated the performance of classification and OOD detection, respectively. We evaluate the classification task by its accuracy and the OOD detection performance by AUROC.

The classification accuracy is described in Table~\ref{table:ImgInDClsfAcc}. The bold font indicates the highest accuracy, and the underline indicates the second highest accuracy. The classification accuracy of synthetic data is about 97\%. The OVADM had the highest accuracy and secondly higher by CE$+$SimCLR. In the image modality, the classification accuracy is improved with self-supervised learning, but OVADM showed lower classification accuracy than CE in all combinations.

\begin{figure*}[t]
\begin{center}
\centerline{\includegraphics[width=0.8\textwidth]{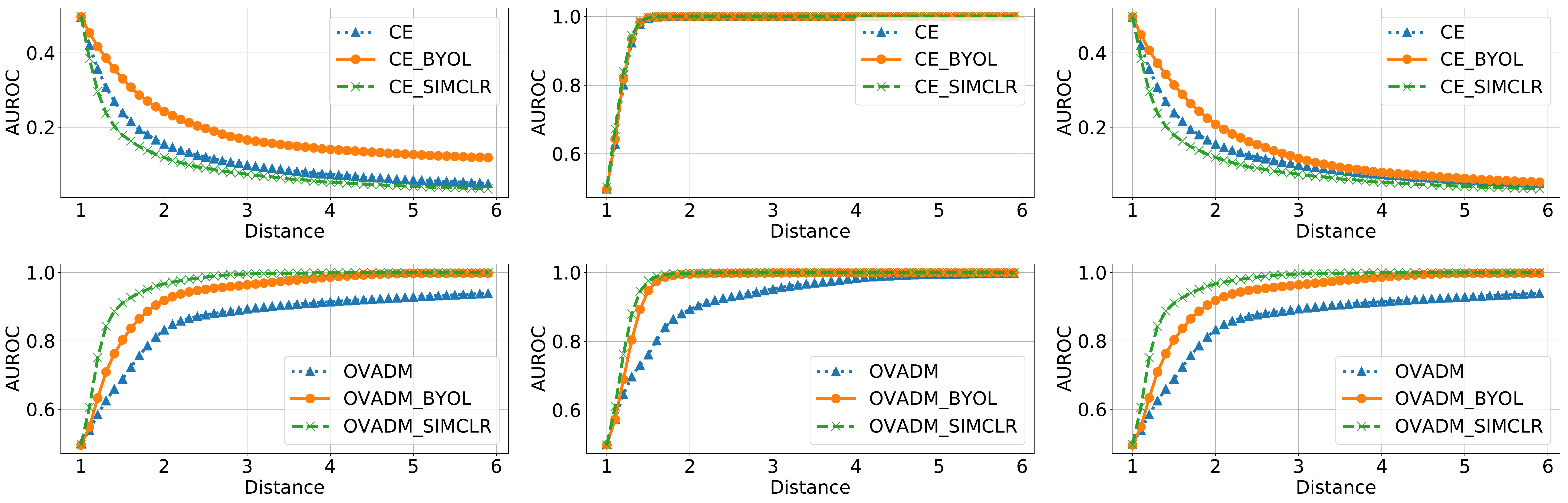}}
\caption{AUROC scores in 2D dataset (Left : Baseline, Middle : MD, Right : ODIN)}
\label{fig:synthetic_auroc}
\end{center}
\vspace{-3mm}
\end{figure*}

\begin{figure}[htb]
    \centering

\begin{subfigure}{0.5\textwidth}
  \includegraphics[width=\linewidth]{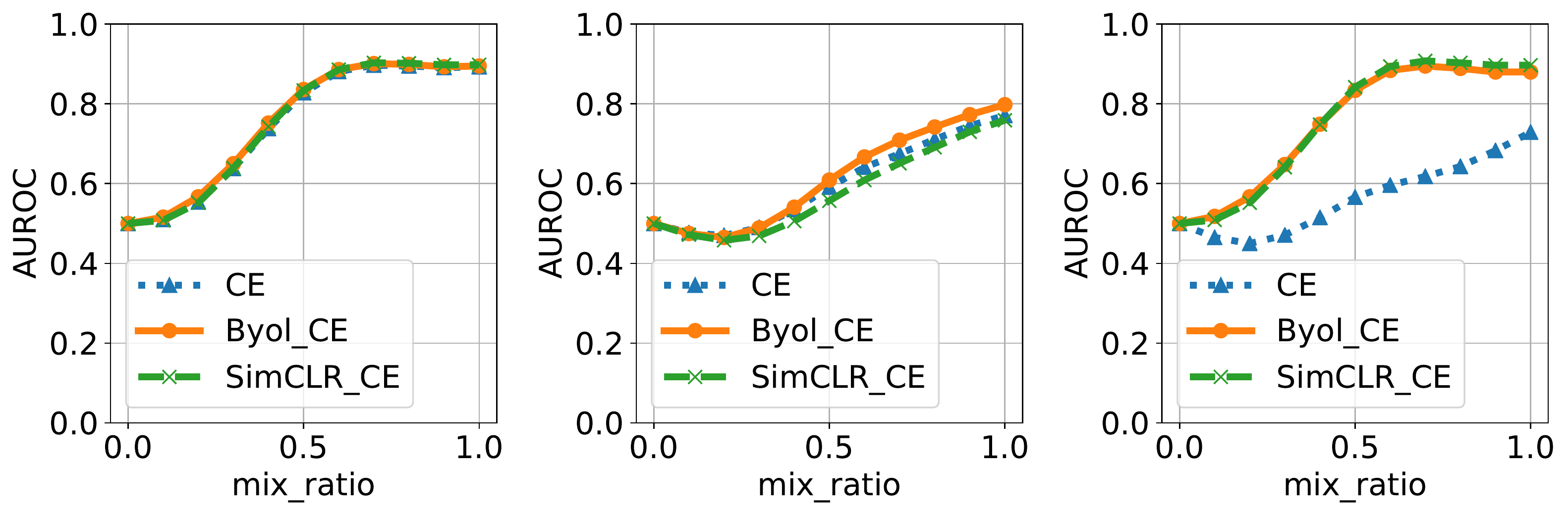}
  \caption{ID:CIFAR10, OOD:CIFAR100}
\end{subfigure}\hfil
\begin{subfigure}{0.5\textwidth}
  \includegraphics[width=\linewidth]{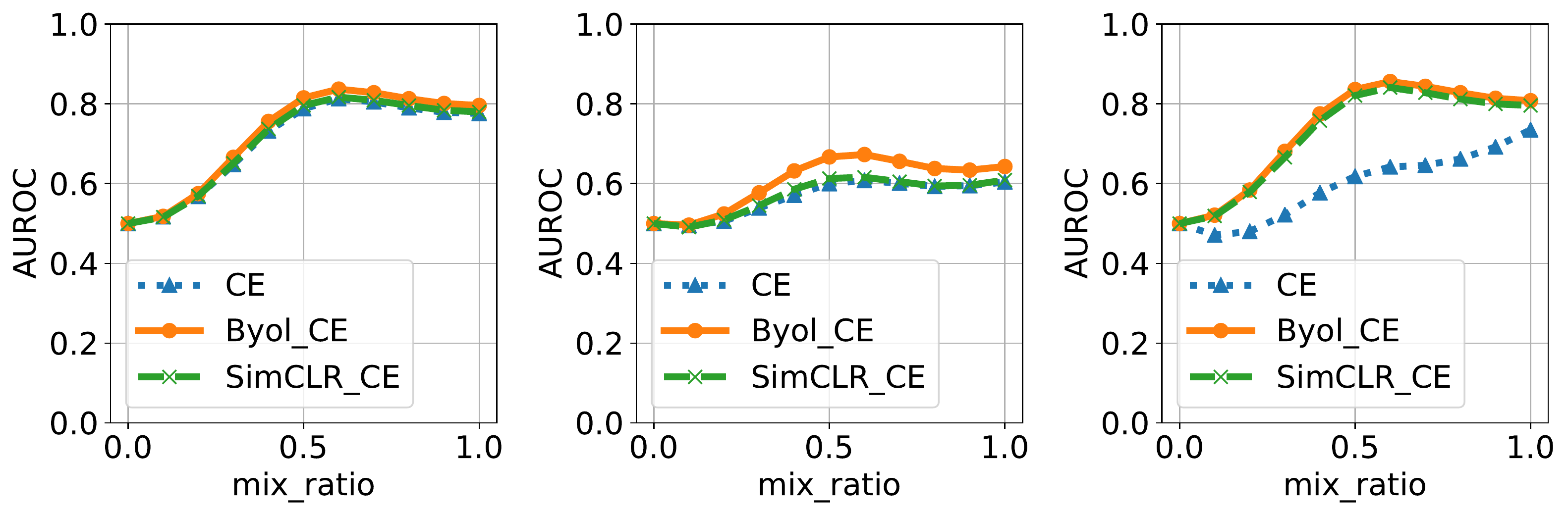}
  \caption{InD:CIFAR100, OOD:CIFAR10}
\end{subfigure}\hfil

\medskip
\begin{subfigure}{0.5\textwidth}
  \includegraphics[width=\linewidth]{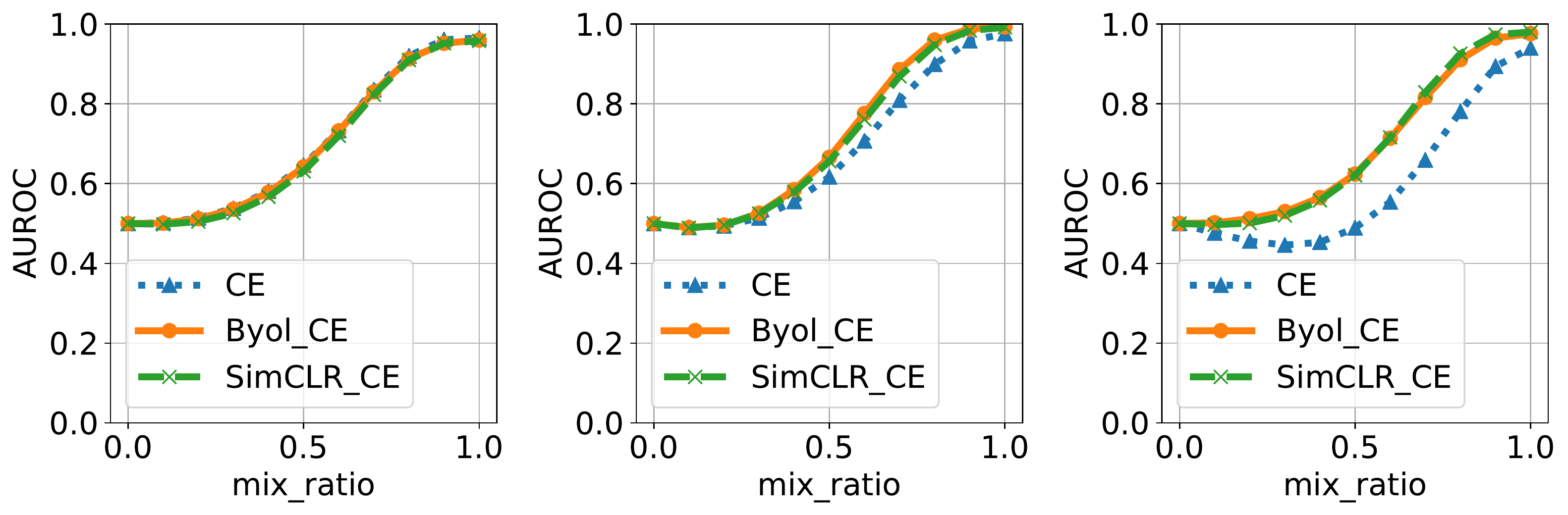}
  \caption{ID:CIFAR10, OOD:SVHN}
\end{subfigure}\hfil 
\begin{subfigure}{0.5\textwidth}
  \includegraphics[width=\linewidth]{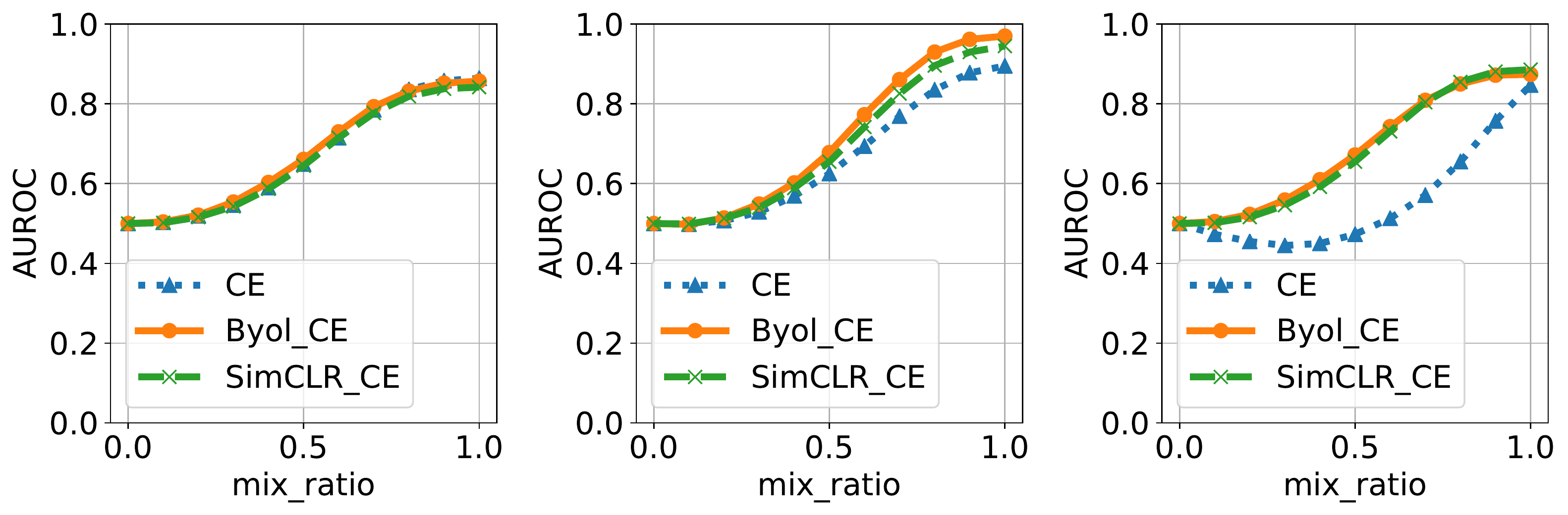}
  \caption{ID:CIFAR100, OOD:SVHN}
\end{subfigure}\hfil 

\caption{OOD performance for images in 4 ID-OOD cases. A mixed ratio of zero means the in-distribution image. (In each case, Left: Baseline, Middle: Mahalanobis distance, Right: ODIN)}
\label{fig:ImgAUROC}
\end{figure}

Even if the models show the same classification accuracy, the OOD detection performance could differ.
The OOD detection performance for the simulated dataset is shown in Fig.~\ref{fig:synthetic_auroc}.
Naturally, distance-based methods (either using OVADM loss or MD OOD score) demonstrate a correct behavior where a larger distance between ID and OOD leads to higher AUROC.
We could also see the AUROCs increase most of the time when SSL is used.
The performance of OOD detection for images is shown in Fig.~\ref{fig:ImgAUROC}.
The AUROC gradually increases as the OOD mixing ratio increases, although the slope and inflection point was different for each OOD dataset.
When self-supervised learning was applied, the inflection point was reached more quickly with a steeper slope than without applying it. 
Comparing the three OOD scoring methods, the AUROC difference between models with and without SSL is not large for the baseline.
On the other hand, ODIN shows a considerable difference in AUROC with SSL, and MD shows a large gap between the model with BYOL and the others.

\section{Discussion}
\label{sec:discussion}
In 2D simulated sample, note that the models using OVADM loss and MD scoring method have the same AUROC performance at $r=5.9$.
But when $r$ is small, the difference increases between models that use SSL and those that do not.
This shows that using a fixed set of OOD data can be ineffective when comparing the OOD detection performances of different methods, which corroborates the need for a controllable OOD evaluation setup.

\begin{wrapfigure}[10]{R}{0.45\textwidth}
    \begin{center}
        \vspace{-8mm}
        \includegraphics[width=0.45\textwidth]{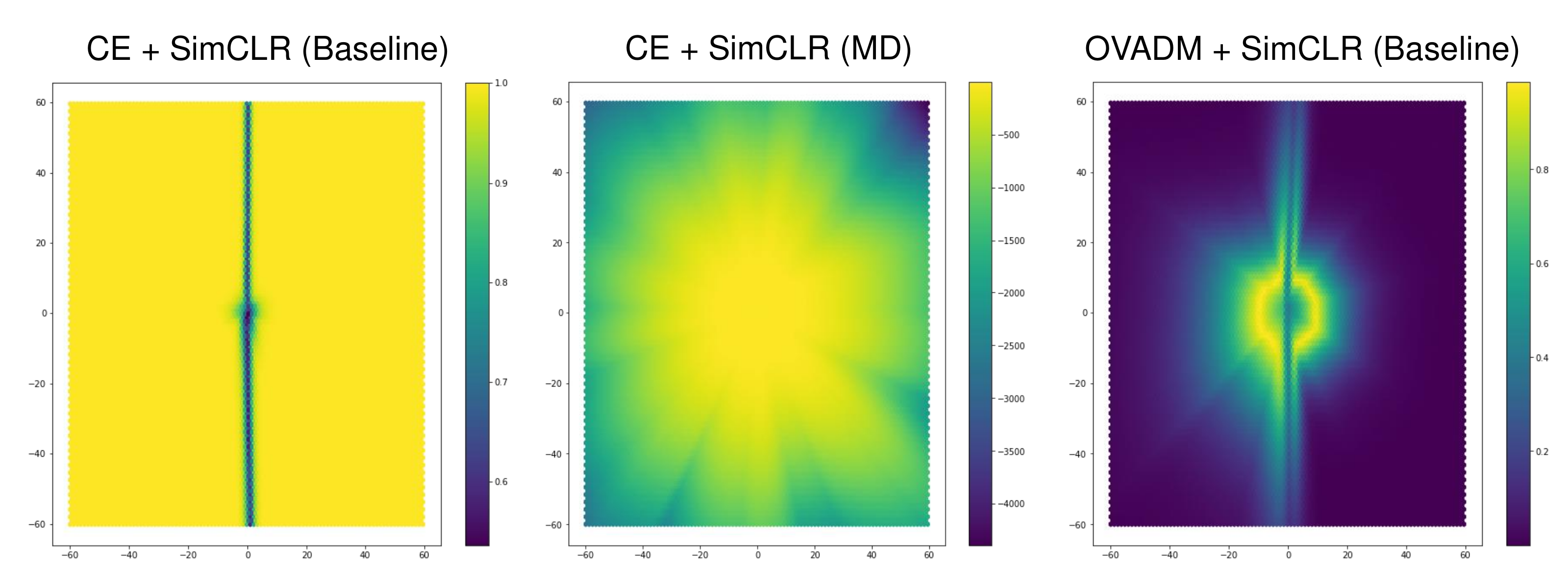}
        \caption{OOD score landscapes in 2D simulated data. (Darker means OOD)}
        \label{fig:synthetic_plane}
    \end{center}
\end{wrapfigure}

Models that use a distance-based method, such as OVADM or MD, increase the value of AUROC as the $r$ increases. However, the AUROC value of models trained without a distance-based method did not increase as the $r$ value increased.
As shown in Fig.~\ref{fig:synthetic_plane}, without the distance-based method (OVADM or MD), a model assigns a high confidence for OOD samples than models using the distance-based methods.

The assumption that the different mixing ratio correlates to the distance between ID and OOD is debatable, but Fig.~\ref{fig:ImgAUROC} shows similar monotonicity as Fig.~\ref{fig:synthetic_auroc} (especially for more distinct dataset pairs CIFAR10-SVHN and CIFAR100-SVHN), indicating the validity of our assumption.
It is notable that SSL not only consistently increases the OOD detection performance, but also guides the model to demonstrate monotonically increasing OOD detection performance as the mixing ratio increases (\textit{i.e.} ID-OOD distance increases). Furthermore, as shown in Fig.\ref{fig:synthetic_auroc} and \ref{fig:ImgAUROC}, the benefit of SSL on OOD detection tended to be more evident on hard samples (\textit{i.e.} close distance between ID and OOD), which is in line with the previous claims  \cite{winkens2020contrastive, tack2020csi}.

\iffalse
\begin{wrapfigure}[10]{R}{0.35\textwidth}
    \begin{center}
        \vspace{-8mm}
        \includegraphics[width=0.3\textwidth]{figs/Max_Accuracy_CIFAR100.png}
        \caption{Classification accuracy according to OOD mixing ratio in computer vision domain}
        \label{fig:ImgClssCIFAR100}
    \end{center}
\end{wrapfigure}
\fi

\section{Conclusions}

\label{sec:conclusion}
In this paper, we proposed the controllable framework for evaluating OOD detection performance.
With this framework, we compared the OOD detection performance and observed the behavior of various methods according to the distance from in-distribution in three different modalities: the 2D synthetic dataset, images, and natural language. As mentioned earlier, the setting of controllable datasets of the computer vision and natural language has limitations, and extension into a more principled and realistic manner using generative models such as VAE\cite{kingma2013auto} or GAN\cite{goodfellow2014generative} can be future work.

\newpage
\section*{Broader Impact}
\label{sec:impact}
We believe this is not applicable to us.

\small
\bibliographystyle{plainnat}
\bibliography{reference}

\newpage
\clearpage
\appendix

\section{Dataset Details}
\label{appendix:dataset}
\textbf{Text}
To test OOD detection in a text-domain, the task of our models is to classify the language of a document into two categories: English and French.
We use IWSLT 2015 dataset \cite{cettolo2015iwslt} which consists of 12 languages.
We randomly select 30,000 examples for training and evaluate the model on 3,000 examples for each language as in-distribution datasets.
Vietnamese documents are used as OOD samples since it is written in alphabet-base characters and linguistically different to English and French.
As can be seen in Fig.~\ref{questiontype}, we construct controllable 6,000 examples by replacing consecutive sentences in in-distribution documents with Vietnamese sentences according to the mixing ratio, where mixing ratio of 1.0 equals pure OOD. 

\begin{figure}[t!]
\begin{subfigure}{0.45\textwidth}
  \includegraphics[width=\linewidth]{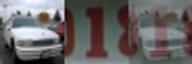}
  \caption{CIFAR10, SVHN, Mixup}
\end{subfigure}\hfil 
\begin{subfigure}{0.45\textwidth}
  \includegraphics[width=\linewidth]{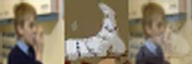}
  \caption{CIFAR100, TinyImagenet, Mixup}
\end{subfigure}\hfil 
\caption{Mixup examples in image domain}
\label{fig:img_mixup_example}
\end{figure}

\begin{figure}[t]
\centering
\begin{tabular}{l|c} Example & Mixup rate \\ \hline
\begin{tabular}[c]{@{}l@{}} 
So what we need to do is to stand up boldly, singly and together, to push \\
governments, to push regional fisheries management organizations, to declare \\
our right to declare certain areas off-limits to high seas fishing, so that the freedom \\
to fish no longer means the freedom to fish anywhere and anytime.
\end{tabular} &
\textbf{0.0} \\ \hline

\begin{tabular}[c]{@{}l@{}} 
\selectlanguage{vietnamese}
\textcolor{red}{Em đã sống những đêm trời có ánh trăng chiếu vàng. Em đã sống những đêm}\\
\selectlanguage{vietnamese}
\textcolor{red}{ngoài kia biển ru bờ cát. Ước gì! anh ở đây giờ này.}\\
our right to declare certain areas off-limits to high seas fishing, so that the freedom \\
to fish no longer means the freedom to fish anywhere and anytime.
\end{tabular} &
\textbf{0.5} \\ \hline

\begin{tabular}[c]{@{}l@{}}
\selectlanguage{vietnamese}
\textcolor{red}{Em đã sống những đêm trời có ánh trăng chiếu vàng. Em đã sống những đêm }\\
\selectlanguage{vietnamese}
\textcolor{red}{ngoài kia biển ru bờ cát. Ước gì! anh ở đây giờ này.}\\
\selectlanguage{vietnamese}
\textcolor{red}{Ước gì! anh cùng em chuyện trò. Cùng nhau nghe sóng xô}\\
\selectlanguage{vietnamese}
\textcolor{red}{ghềnh đá ngàn câu hát yên bình.Em đã biết cô đơn là thế mỗi khi cách xa anh.}\\
\end{tabular} &
\textbf{1.0} \\ \hline

\end{tabular}
\caption{OOD examples in text domain. RED denotes Vietnamese documents used as OOD samples.}
\label{questiontype}
\end{figure}

\section{Implementation Details}
\textbf{2D simulated dataset}
For synthetic-based experiments, we trained a fully-connected network with 2 hidden layers which have 8 units each.
The network was trained for 10 epochs on a batch size of 128 using Adam optimizer.
The learning rate was 0.01.
When utilizing the self-supervised method, we sampled noise values from a normal distribution with a mean of zero and a standard deviation of 0.01. These values are added to the samples to simulate the augmentation of the self-supervision method.

\textbf{Text}
In the text-based expeirment, we took a 12-layer Transformer \cite{vaswani2017transformer} and trained it for the language classification task.
Using stochastic gradient descent with Adam optimizer, we trained the network for 5 epochs with batch size of 16 following the default implementation of HuggingFace \footnote{https://github.com/huggingface/transformers/tree/master/examples/text-classification}.
When using the self-supervised method, unlike the other two modalities, the model was not trained in a multi-task manner.
We used a pre-trained model which was sufficiently trained with self-supervised training objective and we fine-tuned it on the text classification task as \cite{hendrycks2020pretrained}.
We chose XLM-mlm-enfr-1024 model \cite{lample-etal-2019-xlm} as a pre-trained language model considering the in-distribution languages.

\textbf{Image domain}
We used Wide ResNet\cite{zagoruyko2016wide} with depth 28 and width 10 for all the vision-based experiments.
Like the Wide ResNet original paper, we trained the network for 200 epochs with a dropout rate of 0.3 and a batch size of 128.
We used stochastic gradient descent with Nesterov momentum for the optimizer.
The learning rate started at 0.1 and was multiplied by 0.1 at 50\% and 75\% of the training epoch, and the momentum was 0.9.
The data augmentation technique of SimCLR original paper was applied equally to all experiments except for resizing and crop.

\section{Experimental Results from Text}
\begin{figure*}[t]
\begin{center}
\centerline{\includegraphics[width=1.0\textwidth]{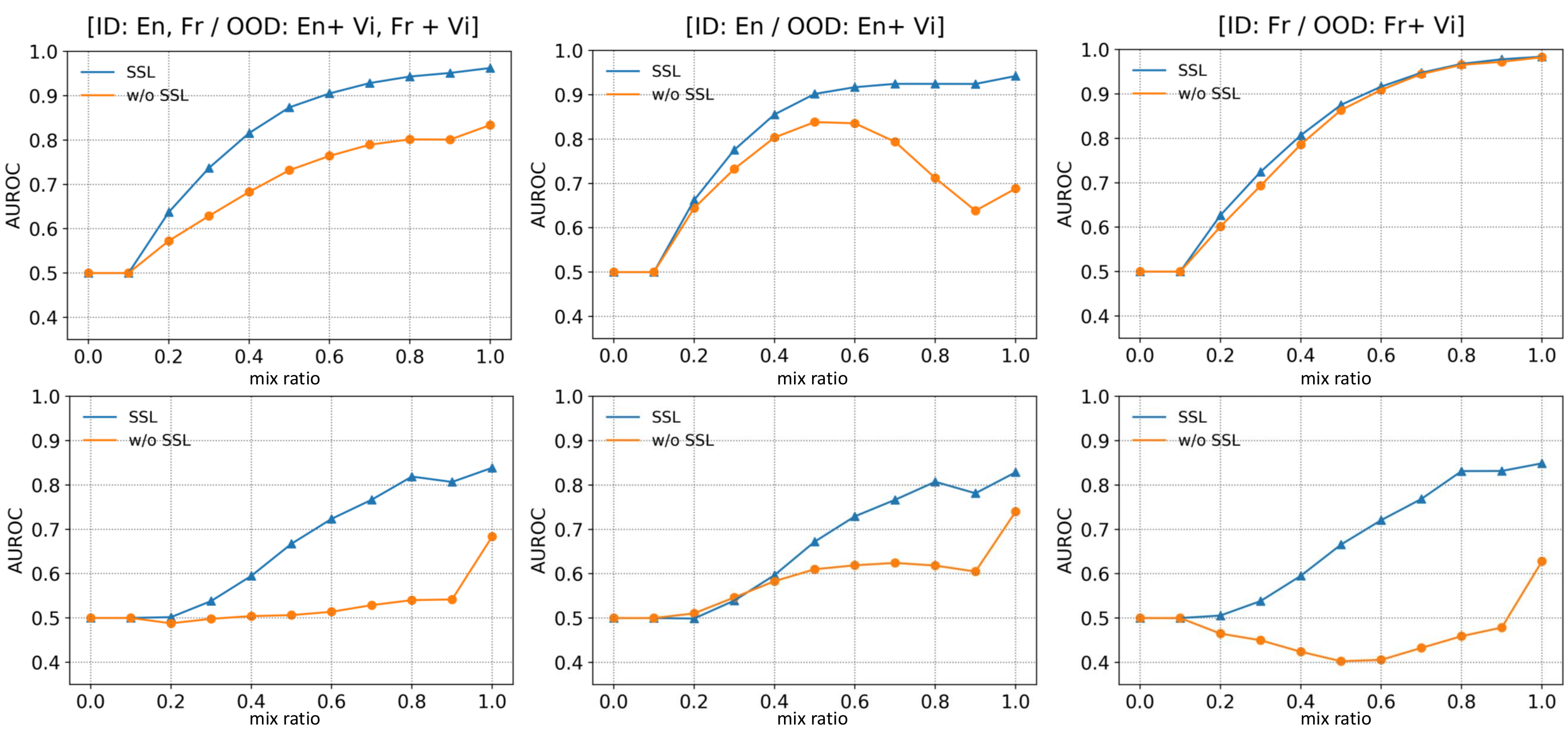}}
\caption{AUROC scores in the text modality using the Baseline (first row) and the Mahalanobis distance (second row). Each column represents the model performance for combined datasets, English only and French only.}
\label{fig:text_results}
\end{center}
\vspace{-3mm}
\end{figure*}

In the text modality, all models were trained to achieve the classification accuracy above 99\%. As shown in the first column of Fig.~\ref{fig:text_results}, SSL-based methods consistently demonstrated better OOD detection performance (\textit{i.e.} higher AUROC scores) compared to the ones without pre-training. 

For further analysis, we separately studied the results of OOD detection for each language (English and French), where the two settings exhibited different behaviors. 
In the English datasets, the one without SSL demonstrated increasing OOD detection performance as the mixing ratio increased (\textit{i.e.,} supposedly, the distance between ID and OOD becoming greater) until the midpoint ($0.6\sim0.7$), but after that, the performance decreased except for the high AUROC when the ratio is $1.0$. Also, in the French datasets, the transformer without SSL using MD showed the opposite behaviors, decreasing OOD detection performance as the mixing ratio increased until the intermediate ratio, but the performance increased after that ratio. It is also notable that the models without SSL using MD in the English datasets demonstrated a similar behavior as (b) in Fig.~\ref{fig:ImgAUROC}.

On the other hand, the SSL-based model showed a consistently increasing OOD detection performance as the mixing ratio increased in all languages.
The overall results from the text modality indicate that the suggested controllable framework for text (\textit{i.e.,} mixing Vietnamese sentences to original sentences) might demonstrate the validity of the SSL in OOD detection. However, we believe the setting of natural language could be extended to a more principled manner, such as using various combinations of datasets with diverse difficulties as ID and OOD.

\begin{figure}[t]
\begin{center}
\begin{subfigure}{0.45\textwidth}
  \includegraphics[width=\linewidth]{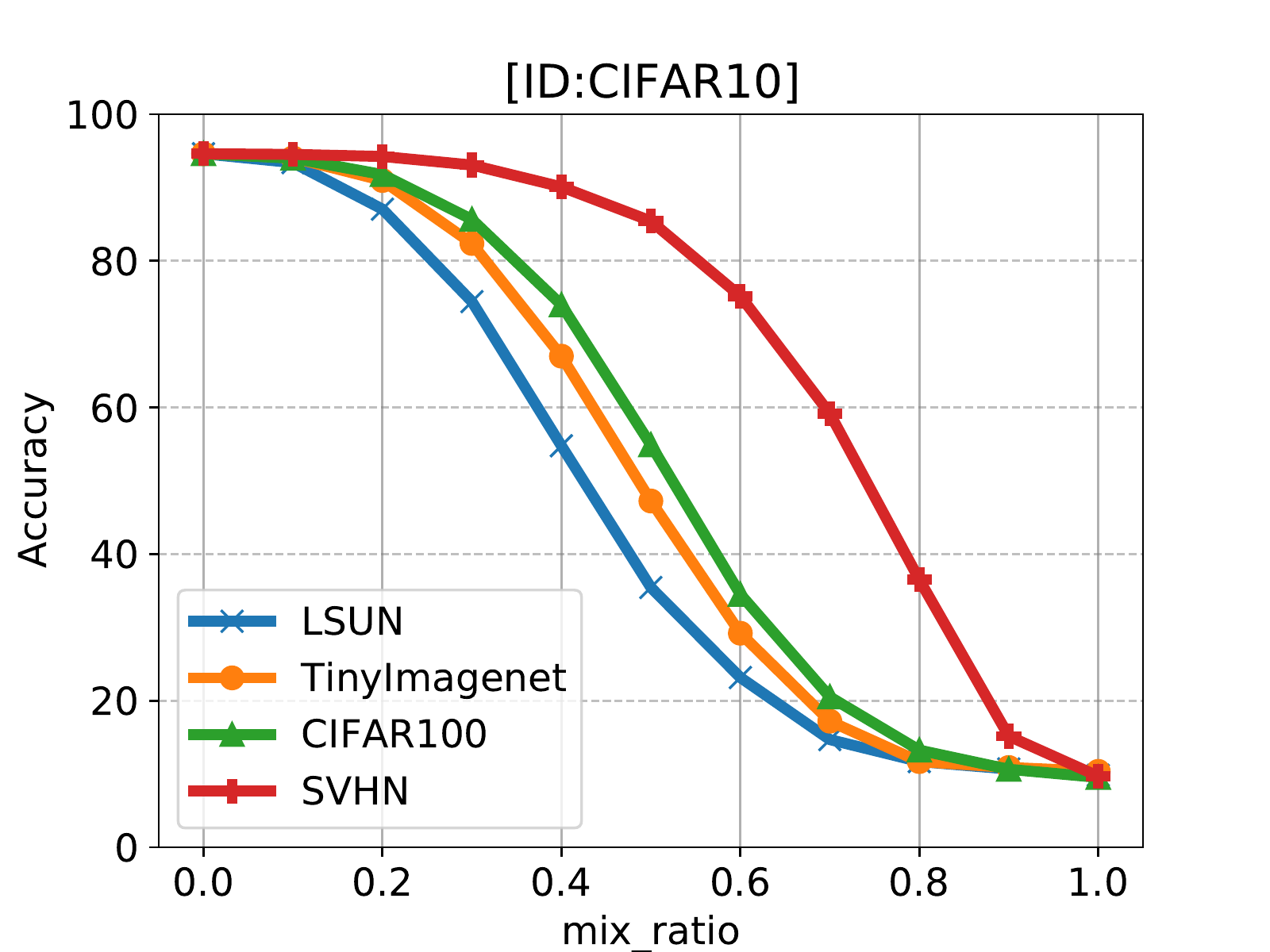}
\end{subfigure}\hfil 
\begin{subfigure}{0.45\textwidth}
  \includegraphics[width=\linewidth]{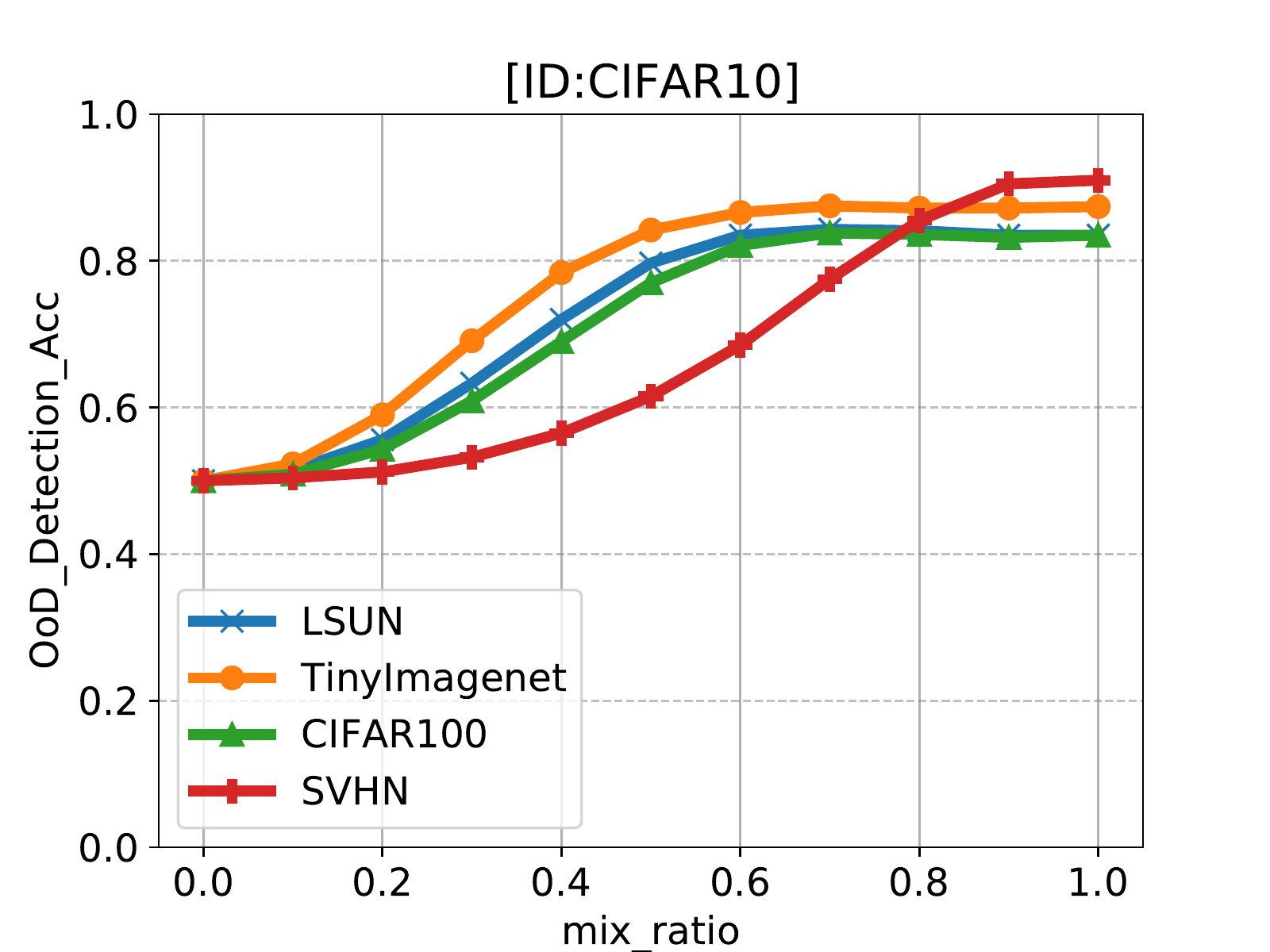}
\end{subfigure}\hfil 
\caption{ID classification accuracy(left) and OOD detection accuracy(right) in image domain when CIFAR10 is utilized as ID}
\label{fig:ImgClassificationAcc}
\end{center}
\end{figure}

\section{Additional Results from Images}
As mentioned earlier, it is assumed that the mixed image gradually approaches from ID to OOD in the image domain, as the mixing ratio increases. As shown in the Fig.~\ref{fig:ImgClassificationAcc}, as the mixing ratio increases, the in-distribution classification accuracy of the mixed image is monotonically decreasing, while the OOD detection accuracy is mostly monotonically increasing.
These results suggest that the proposed method is a valid controllable framework to simulate various difficulty levels of OOD detection.

We found it informative that the speed of change in accuracy depends on the dataset.
When mixing ID with either LSUN, Tiny Imagenet, or CIFAR100 tend to yield a similar effect on the classification accuracy.
On the other hand, when mixing ID with SVHN, the one with most distinct characteristics from ID compared to the other three, the classification accuracy degrades slower as the mixing ratio increases.
Specifically, the classification accuracy stays relatively the same when mixing ID with SVNH, until the mixing ratio reaches 0.3.
When mixing ID with LSUN, on the other hand, the classification accuracy starts to drop as soon as the mixing ratio reaches 0.1.

\begin{figure}[t!]
\begin{subfigure}{0.5\textwidth}
  \includegraphics[width=\linewidth]{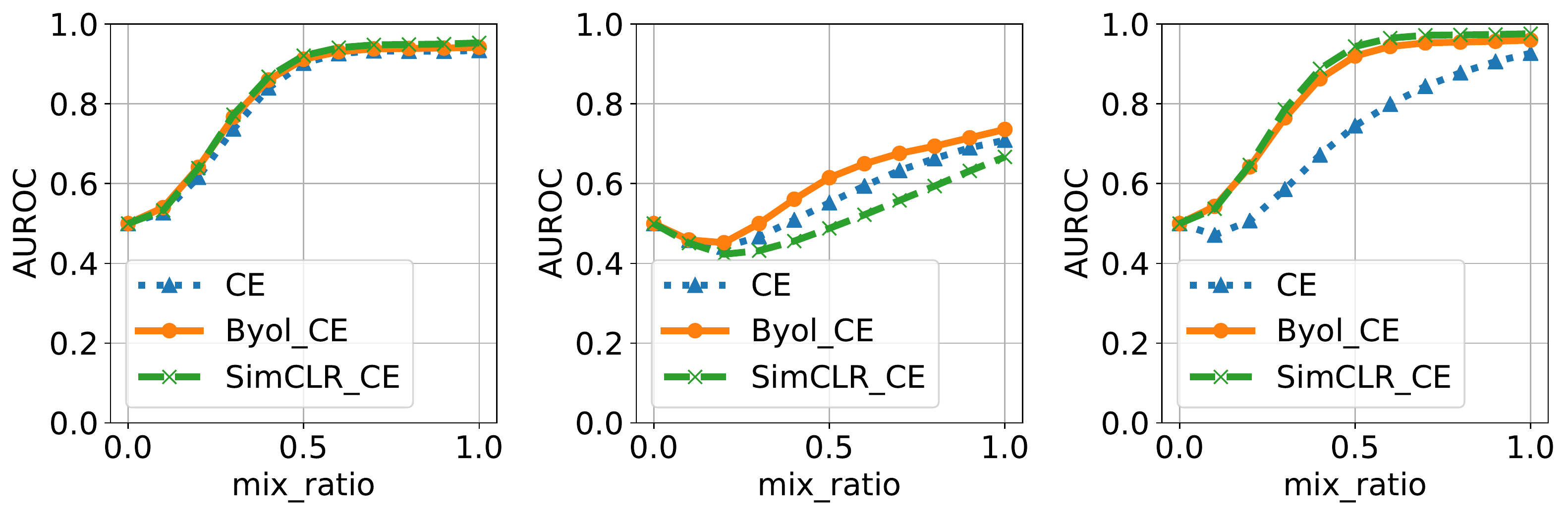}
  \caption{[CE] ID:CIFAR10, OOD:LSUN}
\end{subfigure}\hfil 
\begin{subfigure}{0.5\textwidth}
  \includegraphics[width=\linewidth]{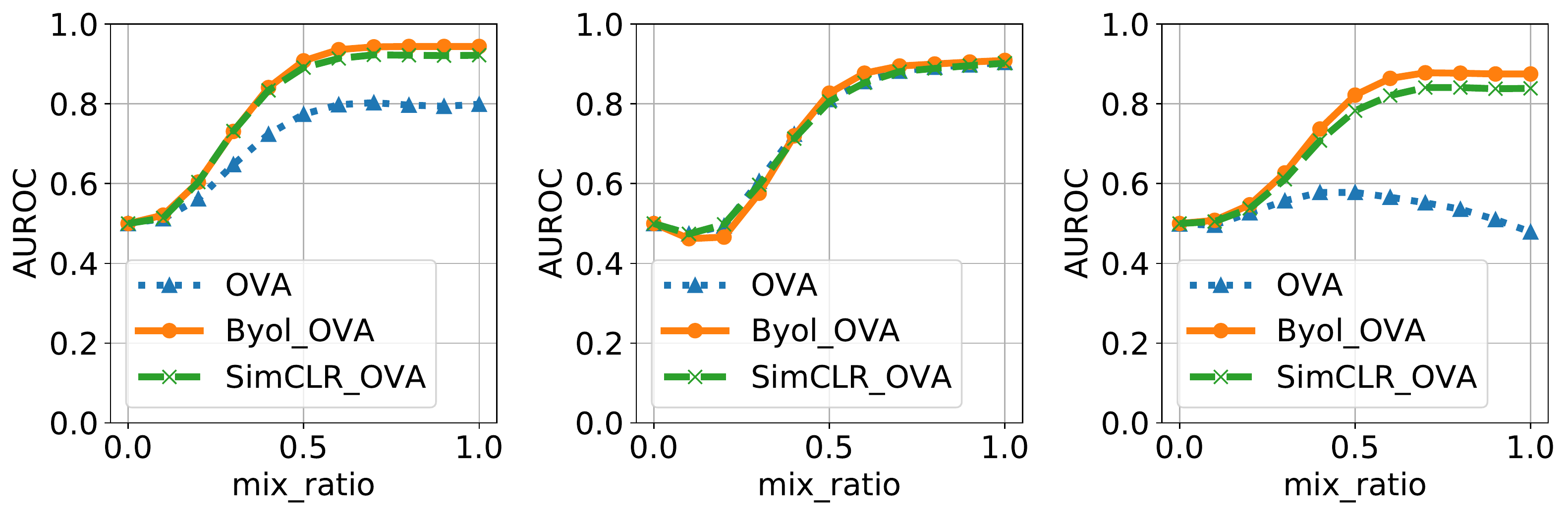}
  \caption{[OVADM] ID:CIFAR10, OOD:LSUN}
\end{subfigure}\hfil 

\medskip
\begin{subfigure}{0.5\textwidth}
  \includegraphics[width=\linewidth]{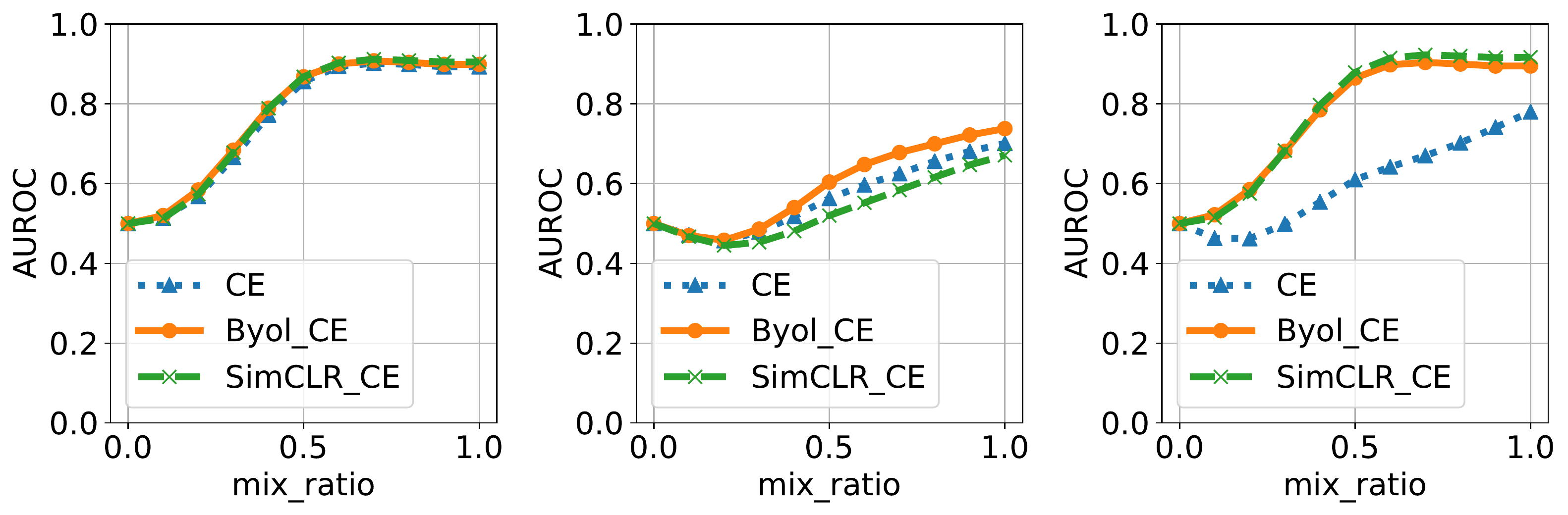}
  \caption{[CE] ID:CIFAR10, OOD:TinyImagenet}
\end{subfigure}\hfil 
\begin{subfigure}{0.5\textwidth}
  \includegraphics[width=\linewidth]{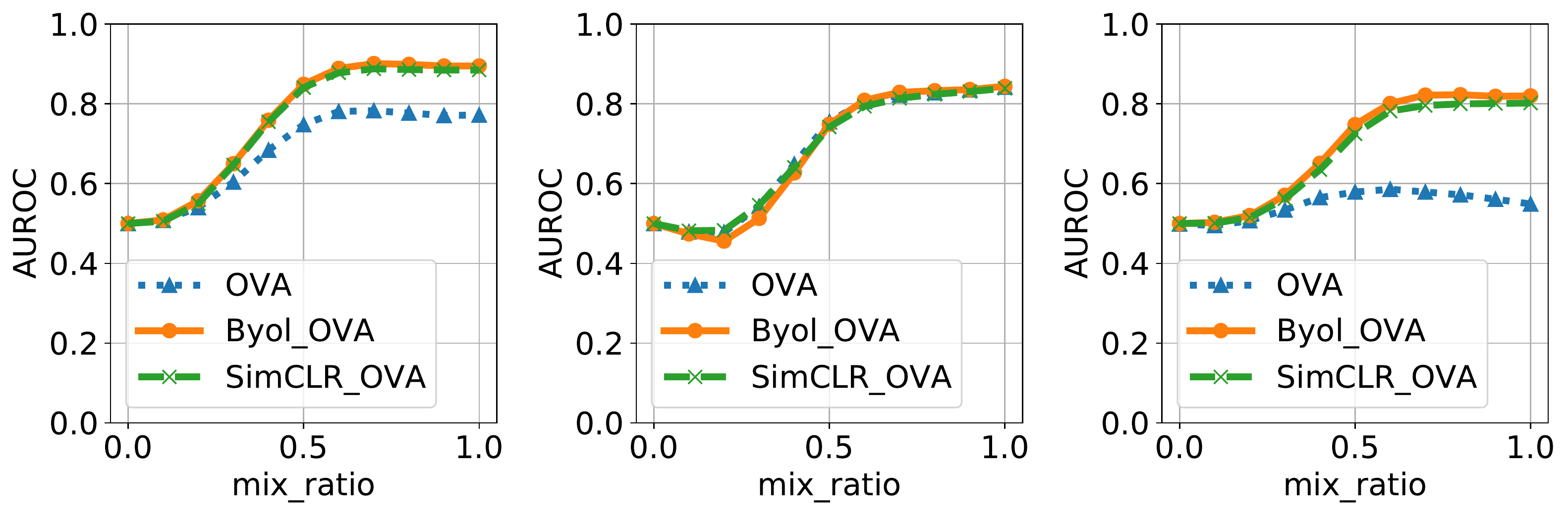}
  \caption{[OVADM] ID:CIFAR10, OOD:TinyImagenet}
\end{subfigure}\hfil 

\medskip
\begin{subfigure}{0.5\textwidth}
  \includegraphics[width=\linewidth]{figs/AUROC_CIFAR10_CIFAR100_CE.pdf}
  \caption{[CE] ID:CIFAR10, OOD:CIFAR100}
\end{subfigure}\hfil 
\begin{subfigure}{0.5\textwidth}
  \includegraphics[width=\linewidth]{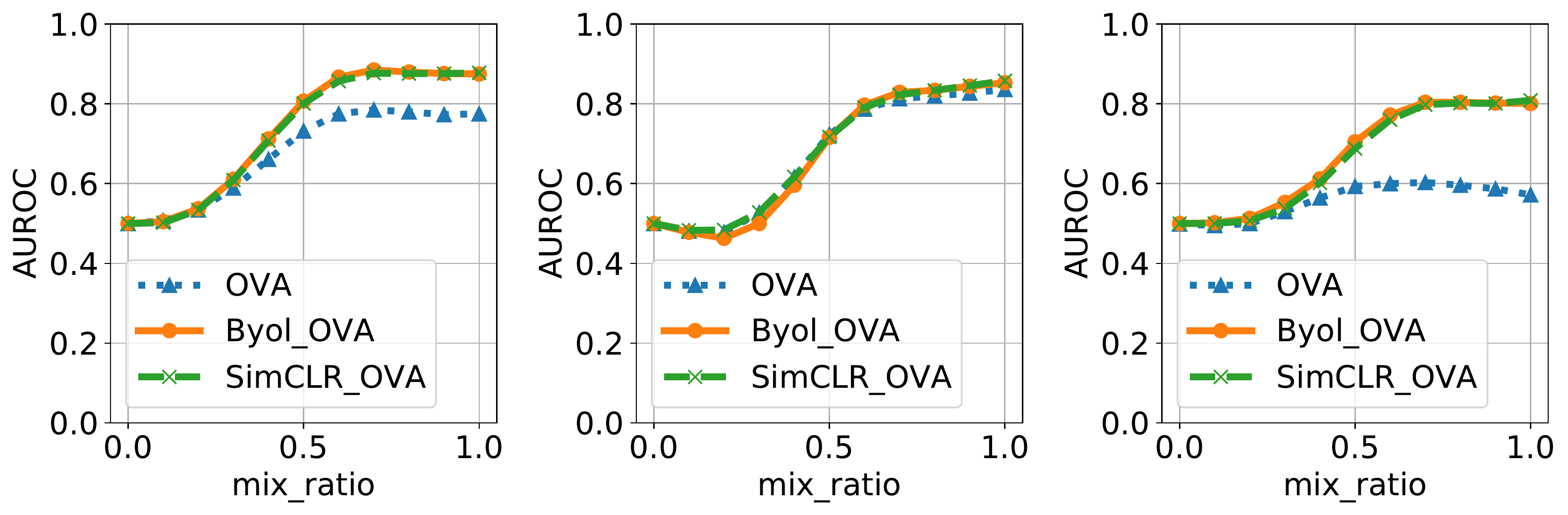}
  \caption{[OVADM] ID:CIFAR10, OOD:CIFAR100}
\end{subfigure}\hfil 

\medskip
\begin{subfigure}{0.5\textwidth}
  \includegraphics[width=\linewidth]{figs/AUROC_CIFAR10_SVHN_CE.pdf}
  \caption{[CE] ID:CIFAR10, OOD:SVHN}
\end{subfigure}\hfil
\begin{subfigure}{0.5\textwidth}
  \includegraphics[width=\linewidth]{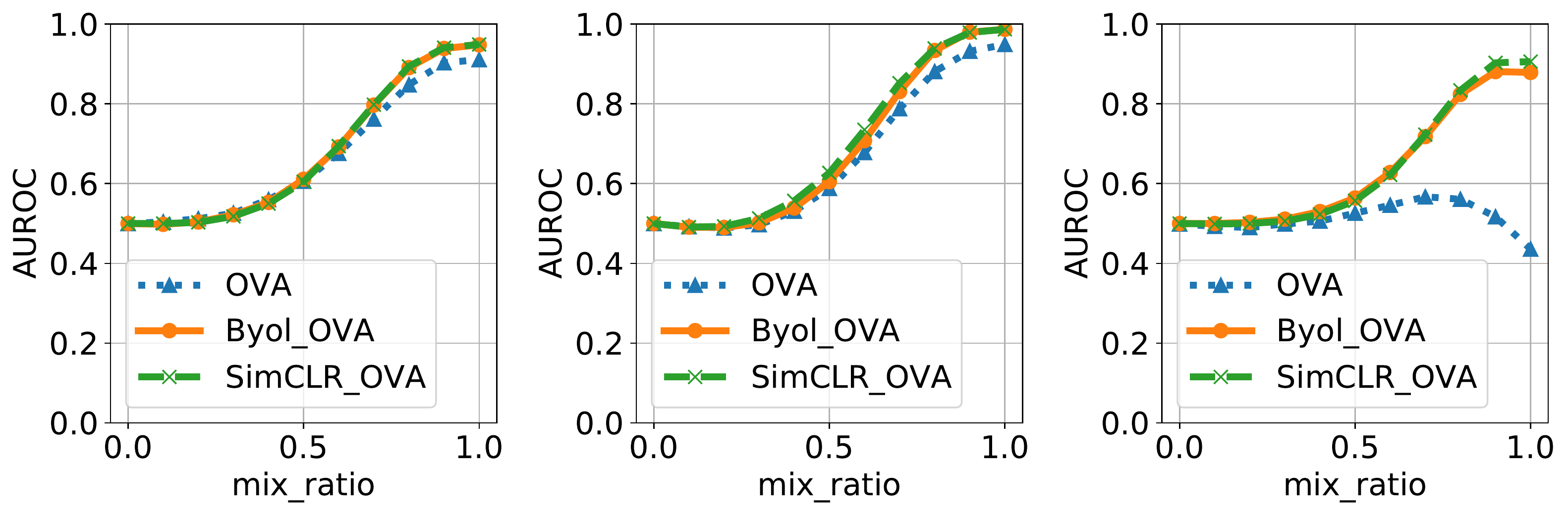}
  \caption{[OVADM] ID:CIFAR10, OOD:SVHN}
\end{subfigure}\hfil 

\caption{Out-of-distribution performance for CIFAR10 in-distribution images. A mixed ratio of zero means the in-distribution image. (a),(c),(e),(g): CE loss is used, (b),(d),(f),(h): OVADM loss is used. (In each case, Left: Baseline, Middle: Mahalanobis distance, Right: ODIN)}
\label{fig:ImgAUROC_CE}
\end{figure}

Fig.~\ref{fig:ImgAUROC_CE} illustrates the AUROC performance of each OOD dataset for CIFAR10 according to the change of mixing ratio. Each row represents a different OOD dataset. The order of these rows was taken from the order of its accuracy curve in Fig.~\ref{fig:ImgClassificationAcc}. The three columns on the left represent the cases where CE loss is applied, and the three columns on the right are the results of applying the OVADM loss.
In addition, the first and fourth columns are the results of using the Baseline, the second and fifth columns are the results of using the MD, and the third and sixth columns are the results of using ODIN.

\begin{figure}[t!]
\begin{subfigure}{0.5\textwidth}
  \includegraphics[width=\linewidth]{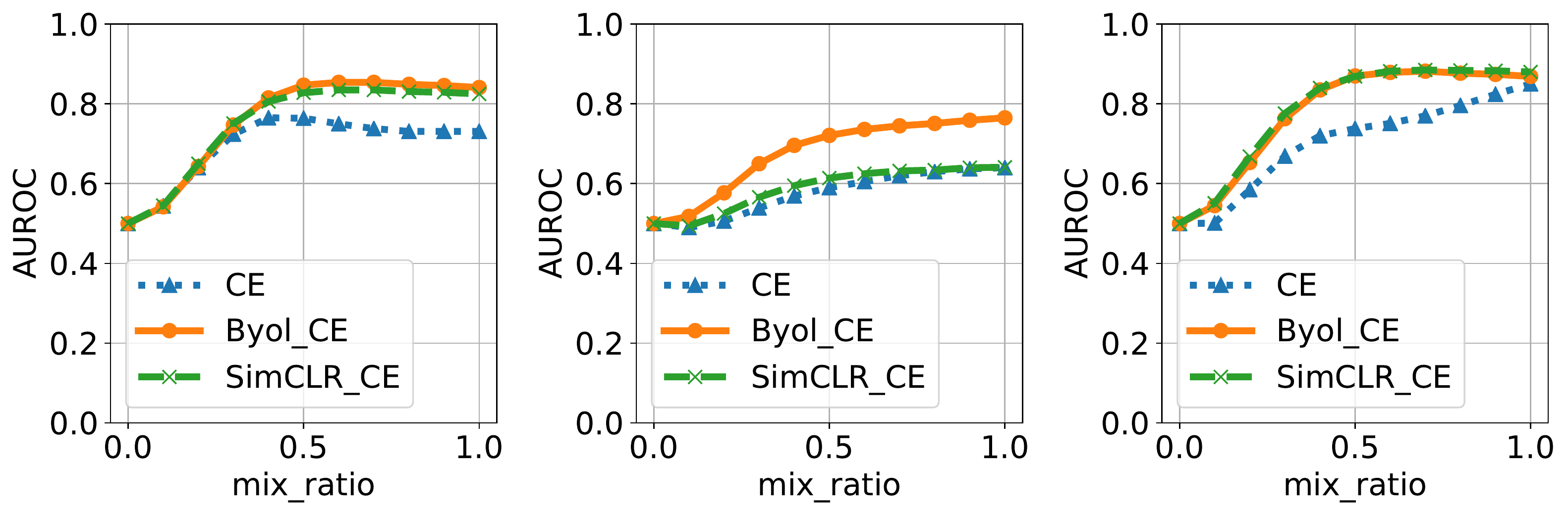}
  \caption{[CE] ID:CIFAR100, OOD:LSUN}
\end{subfigure}\hfil 
\begin{subfigure}{0.5\textwidth}
  \includegraphics[width=\linewidth]{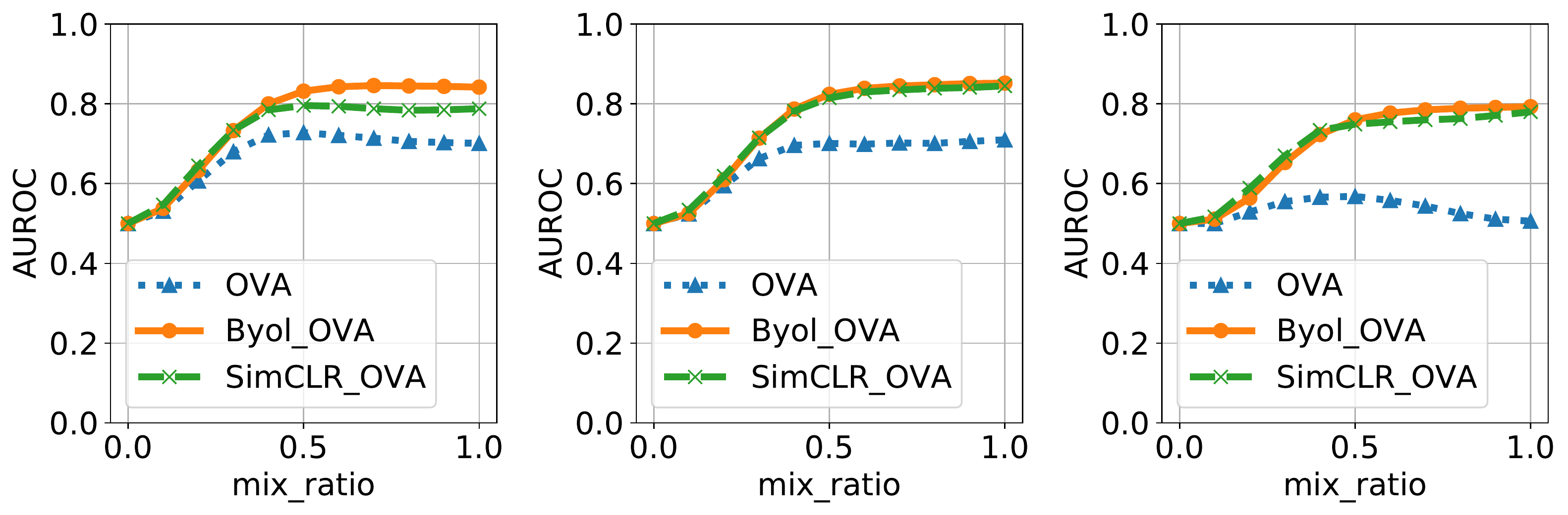}
  \caption{[OVADM] ID:CIFAR100, OOD:LSUN}
\end{subfigure}\hfil

\medskip
\begin{subfigure}{0.5\textwidth}
  \includegraphics[width=\linewidth]{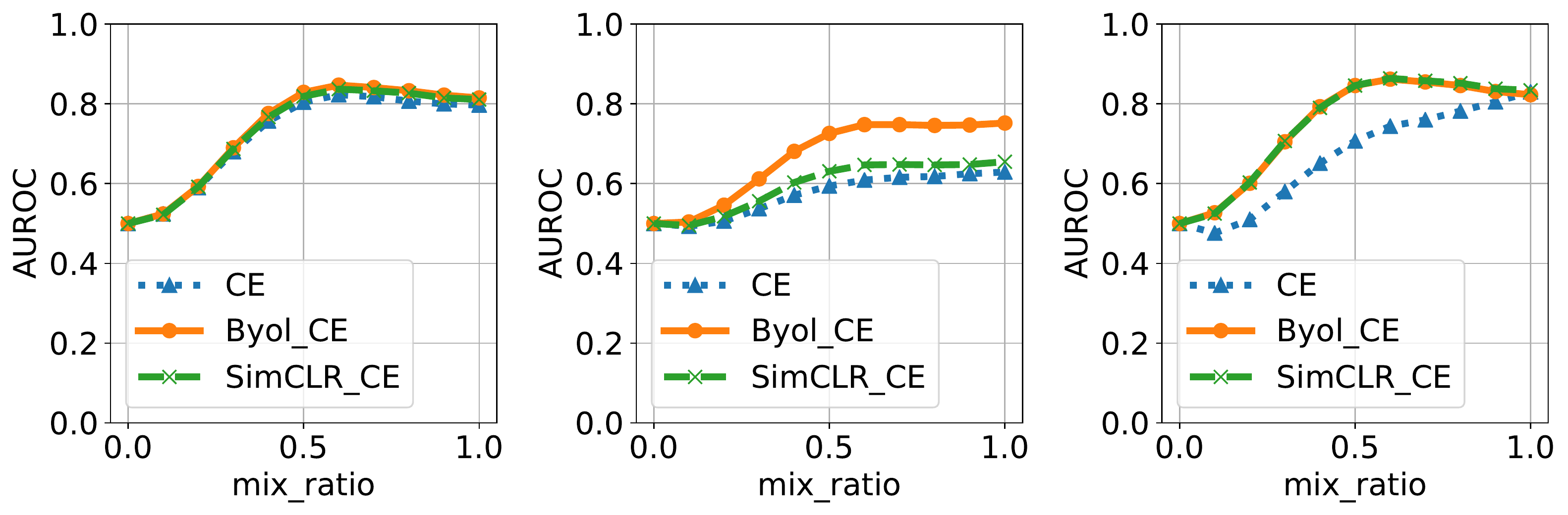}
  \caption{[CE] ID:CIFAR100, OOD:TinyImagenet}
\end{subfigure}\hfil 
\begin{subfigure}{0.5\textwidth}
  \includegraphics[width=\linewidth]{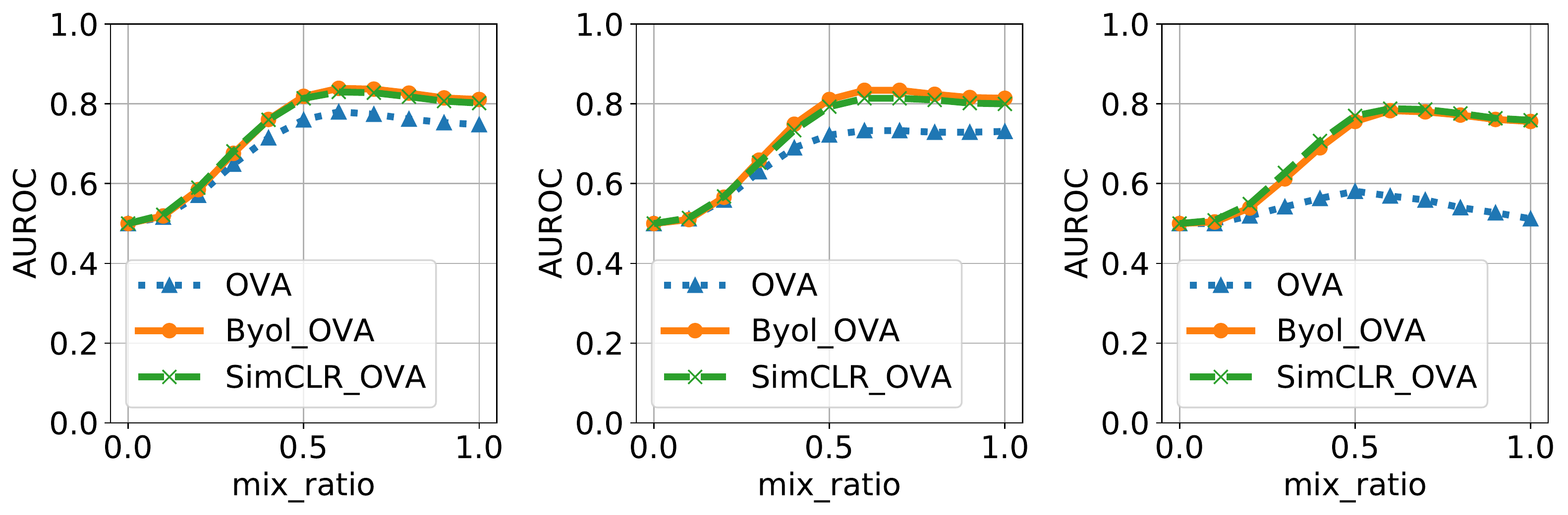}
  \caption{[OVADM] ID:CIFAR100, OOD:TinyImagenet}
\end{subfigure}\hfil 

\medskip
\begin{subfigure}{0.5\textwidth}
  \includegraphics[width=\linewidth]{figs/AUROC_CIFAR100_CIFAR10_CE.pdf}
  \caption{[CE] ID:CIFAR100, OOD:CIFAR10}
\end{subfigure}\hfil 
\begin{subfigure}{0.5\textwidth}
  \includegraphics[width=\linewidth]{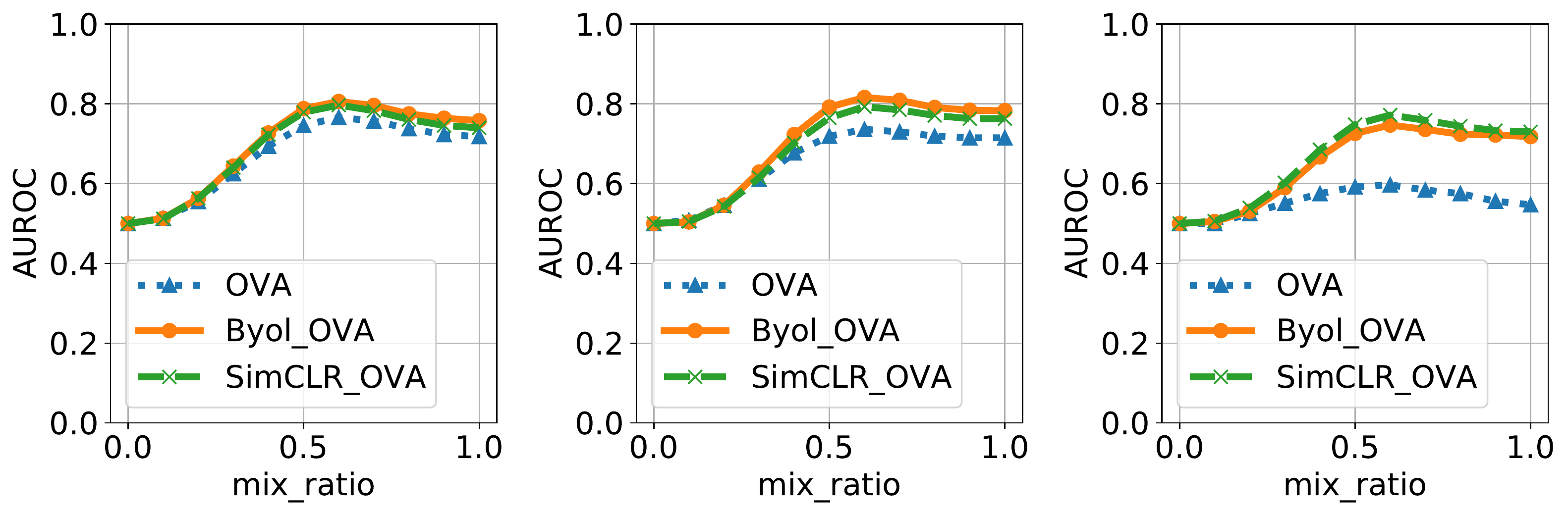}
  \caption{[OVADM] InD:CIFAR100, OOD:CIFAR10}
\end{subfigure}\hfil 

\medskip
\begin{subfigure}{0.5\textwidth}
  \includegraphics[width=\linewidth]{figs/AUROC_CIFAR100_SVHN_CE.pdf}
  \caption{[CE] ID:CIFAR100, OOD:SVHN}
\end{subfigure}\hfil
\begin{subfigure}{0.5\textwidth}
  \includegraphics[width=\linewidth]{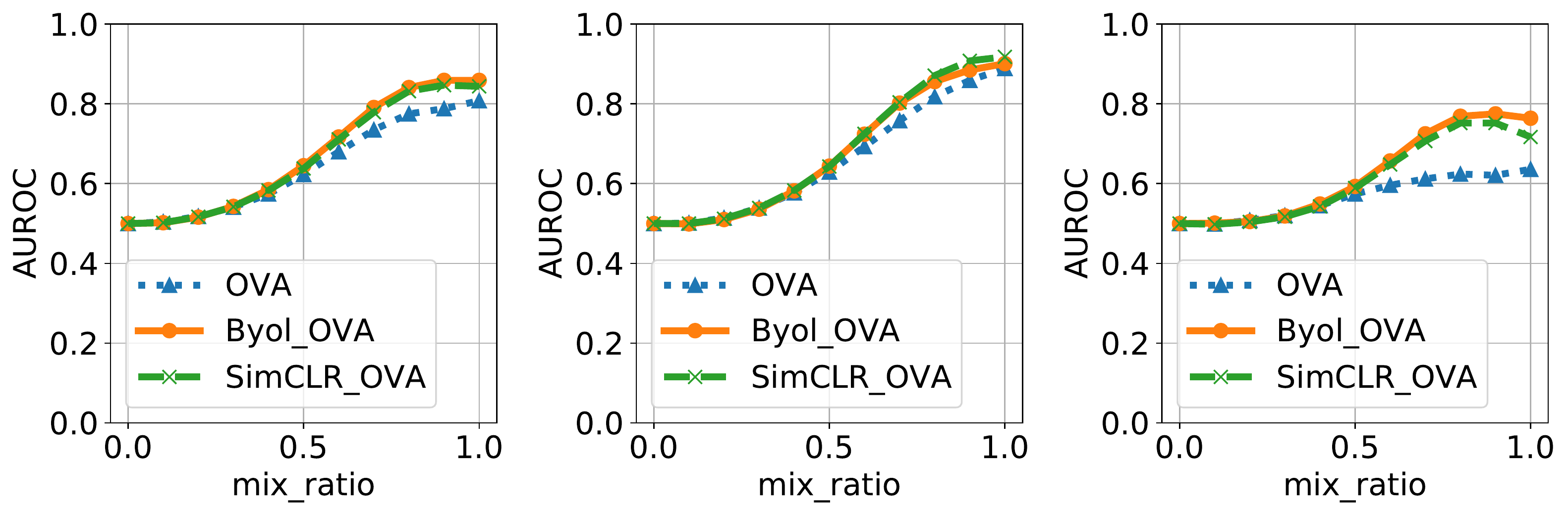}
  \caption{[OVADM] ID:CIFAR100, OOD:SVHN}
\end{subfigure}\hfil

\caption{Out-of-distribution performance for CIFAR100 in-distribution images. A mixed ratio of zero means the in-distribution image. (a),(c),(e),(g): CE loss is used, (b),(d),(f),(h): OVADM loss is used. (In each case, Left: Baseline, Middle: Mahalanobis distance, Right: ODIN)}
\label{fig:ImgAUROC_OVADM}
\end{figure}

First of all, as the mixing ratio increases, the AUROC quickly reaches the inflection point in the order of LSUN, TinyImagenet, CIFAR100, and SVHN, which is the same behavior as the classification accuracy of each datasets.
In other words, the faster the classification accuracy decreases as the mixing ratio increases, the longer the flat section appears in the AUROC plot.  
In addition, when the OOD score was measured by Baseline, the either using SSL or not made little difference when combined with the CE loss.
Strangely, the same behavior was shown when using the OVADM loss, but when the OOD score was measured by MD, not Baseline.
While most of the plot show monotonically increasing AUROC as the mixing ratio increases, the AUROC changes little when using the combination of the OVADM loss and the ODIN method without SSL. 
The similar tendency was also observed when CIFAR100 was used as the ID (Fig.~\ref{fig:ImgAUROC_OVADM}).

\end{document}